\pdfoutput=1

\documentclass[11pt]{article}

\usepackage{acl} %

\usepackage{times}
\usepackage{latexsym}

\usepackage[T1]{fontenc}

\usepackage[utf8]{inputenc}

\usepackage{microtype}

\usepackage{graphicx}
\usepackage[caption=false]{subfig}
\usepackage{amsmath}

\usepackage{booktabs} 
\usepackage{array}
\usepackage{tabularx} 
\usepackage{siunitx}
\usepackage{makecell}
\usepackage{ragged2e}
\usepackage{multirow}
\usepackage{hyperref}

\newcommand{\ignore}[1]{}
\usepackage[show]{notes}
\newcommand{\com}[1]{}

\newcount\Comments  
\newcommand{\kibitz}[2]{\ifnum\Comments=1{\textcolor{#1}{#2}}\fi}

\newcommand{\kg}[1]{\kibitz{red}{[KG:#1]}}

\newcommand{\ab}[1]{\kibitz{brown}{[AB:#1]}}

\title{Detecting Suicide Risk in Online Counseling Services: A Study in a Low-Resource Language}

\ignore{
\author{Amir Bialer \and Daniel Izmaylov \and Avi Segal \and Oren Tsur\\
  Ben-Gurion University of the Negev \\
\AND
   Yossi Levi-Belz \\
   Ruppin academic center \\
\And
  Kobi Gal \\
  Ben-Gurion University of the Negev\\
  University of Edinburgh}
}

\author{Amir Bialer$^\dagger$ \And Daniel Izmaylov$^\dagger$ \And Avi Segal$^\dagger$ \And Oren Tsur$^\dagger$ \AND ~~~~~\{\normalfont {amirbial,zmaylov\}@post.bgu.ac.il ~~~~~~~~~~~~ avisegal@gmail.com ~~~~~~ orentsur@bgu.ac.il} \AND Yossi Levi-Belz$^\ddagger$ \\ yossil@ruppin.ac.il \And Kobi Gal$^{\dagger\mathsection}$ \\ kobig@bgu.ac.il \AND  {\normalfont $^\dagger$Ben-Gurion University of the Negev  $^\ddagger$Ruppin Academic Center $^\mathsection$University of Edinburgh}
}

\begin{document}
\maketitle
\begin{abstract}
With the increased awareness of situations of mental crisis and their societal impact, online services providing emergency support are becoming commonplace in many countries. 
Computational models, trained on discussions between help-seekers and providers, can support suicide prevention by identifying at-risk individuals. However, the lack of domain-specific models, especially in low-resource languages, poses a significant challenge for the automatic detection of suicide risk. We propose a model that combines pre-trained language models (PLM) with a fixed set of manually crafted (and clinically approved) set of suicidal cues, followed by a two-stage fine-tuning process.
Our model achieves 0.91 ROC-AUC and an F2-score of 0.55, significantly outperforming an array of strong baselines even early on in the conversation, which is critical for real-time detection in the field. Moreover, the model performs well across genders and age groups.

\end{abstract}

\section{Introduction}  \label{sec:intro}
 The World Health Organization (WHO) lists suicide as one of the most salient causes of death world-wide, causing more deaths than breast cancer or war \cite{world2019suicide}. 
 With close to $1$ million lives taken directly by suicide every year, and over $25$ million suicide attempts, suicide incurs a lasting impact on families and communities. Identifying individuals at risk ahead of time and providing them with  psychological and medical support is a key step in suicide prevention \cite{joiner2018whether}.

In this paper, we tackle the task of detecting Suicide Ideation (SI). 
Specifically, we aim at the detection of SI of individuals contacting online counseling hot-lines in low-resource and morphologically-rich languages.
We focus on anonymous data from online, text-based (chat), support services. Such services are available in many countries, allowing for confidential and immediate help to those in distress, and play a critical role in suicide prevention  \cite{bantilan2021just,jashinsky2014tracking,joiner2007establishing}. Empirical evidence suggests that at-risk individuals seek help  in close proximity to actual suicide attempts \cite{zalsman2021suicide}, thus  it is critical to identify suicide risks as early as possible during the session. 

There is a growing body of work on assessing suicide risk from English texts (whether in social media posts or from counseling sessions), but there is an acute lack of NLP resources that could be used for detection of suicide risk in other languages \cite{lee2020cross}.
We directly address this gap by focusing on suicide risk detection from online counseling services in Hebrew, which is a low-resource and morphologically-rich language that challenges traditional NLP tasks~\citep{seker2021alephbert}.

Our proposed approach for the SI detection task, called SI-BERT, extends a generic pretrained model with a small set of Out of Vocabulary (OOV) tokens, pretraining the language model on a  masked LM task  and fine tuning for the SI classification task. We further train a logistic regression model, based on a manually crafted lexicon of suicide ideation terms (vetted by domain experts). We then create an Ensemble model by training  SI-BERT together with the lexicon.
  Both SI-BERT and the Ensemble model outperform alternative approaches ranging from W2V (Word2Vec)  embeddings and feed-forward networks to Hebrew psychological lexicon.  Additionally, the Ensemble model achieves 82\% ROC-AUC even when processing only the first few utterances (20\%) of the  help-seeker, suggesting it can be used to enhance early detection of suicide risk when deployed in the field.   We analyze the approach  with respect to different demographics, speaker focus and text truncation, and provide a few examples, qualitatively illustrating the benefits of our model.  Our work goes the first step in helping counselors identify and treat at-risk individuals in real time.


\begin{table*}[h!]
\small
\begin{center}{
  \begin{tabular}
  {cccc}\toprule
    \textit{Paper} &    \textit{Embedding+Model} & \textit{Language}  & \textit{Setting}   \\ \midrule 
   \cite{cheng2017assessing} & LIWC+SVM & Chinese & social media (Weibo) \\
    \cite{allen2019convsent} & LIWC+CNN &English &social media (Reddit) \\
    \cite{matero2019suicide} & BERT without Pretraining & English & social media (Reddit) \\
  \cite{ophir2020deep} & ELMO+Questionnaires+ANN& English& social media (Facebook) \\
      \cite{lee2020cross} & W2V+LSTM+Lexicons & Korean& social media (Naver Cafe) \\
      \cite{bantilan2021just} & TF-IDF+XGBoost & English& phone counseling \\
    \cite{xu2021detecting} & Knowledge Graph+W2V+LSTM & Chinese & online counseling 
 \\ \bottomrule
  \end{tabular}
    \caption{Sample of relevant approaches used for  suicide risk classification from text.  }
     \label{tab:Literature_Review}

     }\end{center}

\end{table*}

\section{Related Work} \label{sec:Related} 
Our work extends past approaches to suicide risk detection in texts as well as NLP classification tasks in low-resource languages. 
For a  systematic review of the use of machine learning for suicide risk detection from text we refer the reader to  \citet{ji2020suicidal} and to \cite{bernert2020artificial}. For a comprehensive survey of the application of the BERT architecture in different scenarios we refer the reader to \cite{rogers-etal-2020-primer}.
In the remainder of this section we briefly survey recent works in each of these fields.

\paragraph{Detection of suicide risk} There are limited works in suicide detection in conversations between help-seekers and counselors.  \citet{xu2021detecting} used a classifier based on a knowledge graph of logical relationships of events related to suicide ideation. They combined this graph with  Word2Vec embeddings to detect suicide risk in an online counseling service in Hong Kong. Their model achieved  $81.5$\% ROC-AUC for suicide risk detection. \citet{bantilan2021just} combined TF-IDF embeddings with  an  XGBoost model in transcribed phone calls from a counseling service in English.  Their approach achieved a 73\% ROC-AUC performance in the 
 phone call based counseling. 
 None  of these approaches addressed early detection, and are outperformed by our own approach in terms of ROC-AUC.

There is a body of work on suicide risk detection for English text   from social media posts \cite{guntuku2017detecting,de2013predicting,zirikly2019clpsych}. Online counseling chats are quite different from such settings in that they often include complete conversations between help-seekers and counselors, rather than single utterances, and  exhibit temporal and mental-state dynamics.

 \citet{ophir2020deep}, used ANNs
 to predict at-risk individuals from  Facebook posts and psychological questionnaires.
 ~\citet{matero2019suicide} achieved top results in a  suicide ideation detection task in social media~\cite{zirikly2019clpsych} by adapting a BERT model to process input from Twitter and Reddit posts.  They found that at-risk individuals  use a distinct vocabulary in comparison to the rest of the population. 
 \citet{lee2020cross} tackled suicide ideation detection in social media posts in Korean which they describe as a low resource language. In their work they claim domain lexicons are highly beneficial for the task when available. Their classifying model is based on word embeddings, lexicons, attention, and LSTM.
These works inspired our approach to combine expert based lexicons with the language model.  

For convenience, the main approaches and the settings on which they were developed are summarized in \autoref{tab:Literature_Review}.

\paragraph{Using Transformers for low-resource and morphologically rich languages} Deep neural architectures, especially the Transformer, require massive corpora for adequate training. These pretrained models are then fine-tuned for specific classification tasks, e.g., \cite{devlin2018bert,sun2019fine,pierse2020aligning,gururangan2020don}. However, fine tuning is suboptimal in the cases where the domain of the classification task is unique, especially in low resource and morphologically-rich languages \cite{klein-tsarfaty-2020-getting,seker2021alephbert,nzeyimana2022kinyabert}. Our approach tackles those issues.

\paragraph{Hebrew NLP} There is an increasing number of Hebrew tools available for modeling NLP tasks.  \citet{shapira2021hebrew}  released a Hebrew Psychological Lexicons (HPL) that contains 30\% of the terms that exist in LIWC, while also containing unique psychological terms that can help detect psychological aspects such as emotional state. 
Other tools include generic Hebrew BERT models such as HeBERT ~\cite{chriqui2021heBERT} and AlephBERT \cite{seker2021alephbert}.  AlephBERT was trained on a larger dataset and achieved better results than HeBERT, making it our PLM of choice. 



\section{The Dataset} \label{sec:dataset}
 Sahar (\url{https://sahar.org.il}), Hebrew acronym for ``Aid and Attention Online'',
is the leading chatline in Israel, focusing on suicide prevention, and emotional distress relief. Relieving the emotional distress of help-seekers is a crucial step in the process of suicide prevention \cite{overholser1997emotional, suris1996chronic}. The organization handles more than $10,000$ chat sessions a year, and these numbers have increased significantly during the COVID-19 pandemic \cite{zalsman2021suicide}.

Sahar's counselors are volunteers over 24 years old that completed a special training program by licensed clinicians. 
They are trained to use a special support language that is based  on the conversation dynamics. 
During shifts, there are also therapists on duty who monitor the conversations and provide professional support if needed to the counselors. 
At the end of each session, the counselor is asked to summarize the 
conversation and to complete a short survey. 
A conversation is defined as exhibiting SI if the counselor answered ``Yes'' to the   question  ``Did the subject of suicide come up in the conversation?''

\paragraph{The Sahar corpus} The Sahar corpus contains 44,506 chat sessions which took place in the span of  five years (2017-January 2022). Of these,  17,564 are labeled with a True/False to designate whether the chat exhibited SI.  Seventeen percent $(3097/17564)$  of the labeled sessions are flagged with a positive SI label.   For the remainder of this paper, the term Sahar dataset will refer to the 17,564 labeled sessions in the corpus.
A session includes the utterances generated by the help-seekers and the counselor, delimited by a separating token.  \autoref{tab:data_table} shows general statistics related to our dataset.

\begin{table}[h]
\small
    \begin{center}{
        \begin{tabular}{cc}
        \\
        Dataset statistics
        \\
        \hline
        \rule{0pt}{10pt}
        Total Num.  Sessions  & $44,506$ \\
        Num.  Labeled Sessions  & $17,564$
        \\
        SI positive label ratio  & $17\%$
        \\
        Mean(Median) number of tokens in a chat & $617(566)$
        \\
        \hline
        \end{tabular}
        \caption{General statistics for Sahar Corpus}
        \label{tab:data_table}
    }\end{center}
\end{table}

\begin{figure}[t]
    \centering
     \includegraphics[width=0.48\textwidth]{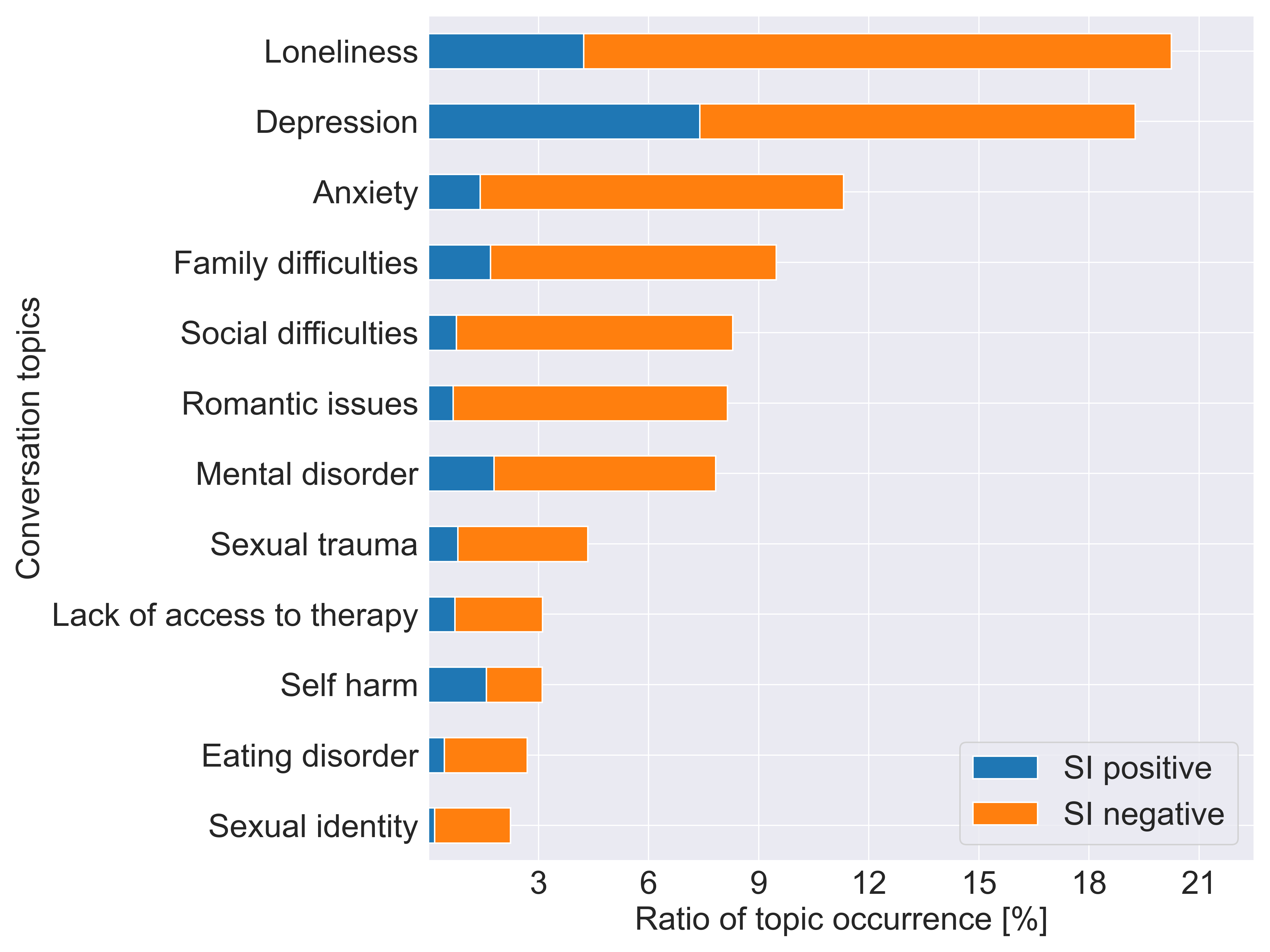}
      \captionsetup{width=.9\textwidth}
     \caption{Conversation topics distribution in Sahar corpus 
   }
    \label{fig:topics}
\end{figure}

 Beyond the SI label, counselors are requested to select the prominent topics in each conversation (one to three topics chosen from a predefined list). 
\autoref{fig:topics} shows the discussed topics distribution in the data.  As shown in the figure,  loneliness and depression are the most common topics discussed in Sahar platform and a large share of those conversation exhibits positive SI. These findings coincide with psychology literature which depicts depression \cite{gijzen2021suicide,moitra2021estimating} and loneliness \cite{mcclelland2020loneliness} as powerful predictors for suicide ideation. An interesting observation from  \autoref{fig:topics} is, that SI is prevalent across all of the reported topics, and is not exclusive to a certain topic.

Unfortunately, we are not able to share the Sahar corpus due to it's sensitive and private content. Nor are we able to share the trained language model, since it was shown language models can be manipulated to reveal training data (e.g \cite{carlini2021extracting}). We do provide however  a  \href{https://github.com/AmirBialer/Coling_2022_Early-detection-of-Suicide-Risk-in-Online-Counseling-Services}{repository} with the experiments' code and lexicons used in this paper to support transparency and reproducibility. 

\begin{figure*}
    \centering
     \includegraphics[width=0.98\textwidth]{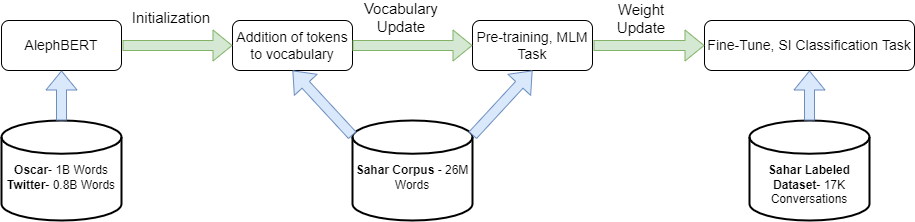}
      \captionsetup{width=.9\textwidth}
     \caption{SI-BERT Architecture}
    \label{fig:pipeline}
\end{figure*}

\section{Computational  Approach} \label{sec:meth}
 Our approach is based on an Ensemble method  
that extends a generic PLM to the SI domain, and leverages it with   a fixed set of manually crafted (and clinically approved) set of suicidal cues.  
We expand on each component of the Ensemble model in turn. 

\subsection{The SI-BERT Classifier}
\label{subsec:si-bert}
SI-BERT is a model designed and configured for SI detection in online Hebrew chats. It utilizes AlephBERT \cite{seker2021alephbert}, which is the best performing Hebrew BERT model to date and is publicly available. 
SI-BERT  augments the generic AlephBERT model by (a) adding domain-specific tokens to the vocabulary; (b)  pretraining on a masked language model (MLM) task over the Sahar corpus; and (c) fine-tuning for the SI classification task. 
The SI-BERT architecture is illustrated in \autoref{fig:pipeline}.  The term `SI-BERT'  at each step  refers to the model obtained by the previous step. In the remainder of this section we expand on each of these steps.

\paragraph{Adding Domain Specific Tokens} 
It is well established that enriching the vocabulary of pretrained models with Domain Specific Tokens (DST) improves performance for domain specific tasks~\cite{tai2020exbert,beltagy2019scibert,honda-etal-2021-removing}.

We consider words which are OOV to be domain specific tokens, since by definition they exist in the domain corpus and not in the language model's vocabulary. The language model's vocabulary was constructed by taking the most common tokens from it's training corpus.
Manual examination of the most frequent DST finds that many of these tokens are highly related to suicide and mental distress e.g, ``suicidal'', ``desperate'', ``depressed'', ``abandoned'' (translated from Hebrew). We hypothesize that adding a relatively small number of domain specific tokens will improve performance for suicide ideation classification task.

Therefore, the DST list we used is the $\delta$  most frequently appearing words in the Sahar corpus that were not in the vocabulary of the pretrained model. We added this list to SI-BERT's vocabulary, changing it's size from $|V|$ to $|V|+\delta$. We set $\delta=1000$ based on performance on a held-out validation set. Analyzing the number of OOV token in Sahar corpus, we find 5\% of the words to be OOV (before the addition of the DST list). After adding the DST to the vocabulary we observed a decrease of 20\% in the number of OOV words in Sahar corpus.

\paragraph{Pretraining with MLM task}
In this step, we pretrain SI-BERT with a Masked Language Model (MLM) task. MLM is an unsupervised task in which a share of the  tokens  in each utterance is masked,   and SI-BERT  is trained to predict the masked words.  This task has been shown to improve the performance of BERT and other language models performance for downstream tasks \cite{gururangan2020don, pierse2020aligning}.
The training for the MLM task was  conducted for 200 epochs on the complete Sahar  corpus. An additional advantage of the MLM task is that it retrains the 
weights of the SI-BERT following the additions of tokens in the previous step \cite{tai2020exbert}.

\paragraph{Fine-Tuning}
In this step we fine-tune SI-BERT for the SI classification task.  
We add a binary classification head to SI-BERT. The classification head is a neural network layer which consists of two neurons with a softmax activation (binary classification). We compile the model with a cross entropy loss. We fine tune our model with the labeled dataset described in \autoref{sec:dataset}. 
The BERT model is designed to process 512 tokens. We fine tune the model with the help-seeker's text
and used the first 512 tokens of each session as input to the model. For a discussion and justification of these decisions, see \autoref{subsec:trunc}. In practice,  21\% of the sessions were truncated when inputted to the model.

\subsection{Suicide Ideation Lexicon  } \label{ssec:lex}
We extracted 200 randomly selected positive SI sessions from the  Sahar dataset 
and constructed a list of phrases that explicitly mention suicide ideation. The list contains 67 phrases  such as: ``suicide'',
``cut wrists'', ``want to die'' and other variations.  

This list was vetted by psychologists with expertise in suicide ideation. 
The set of 200 positive sessions was removed from the test sets used in the evaluation. 
 Each session is mapped to a vector of length 67, where the $i$th element in the vector is the number of occurrences of phrase $i$ in a given session. The vectors are scaled to [0,1] range and fed into a logistic regression model.
 
\subsection{Ensemble Model}
\label{subsec:ensemble}
The Ensemble model combines SI-BERT and the lexicon by feeding their predictions to a fully connected layer activated with a sigmoid function. 

\section{Evaluation}
We compared the Ensemble model to the baseline models described below.  The input to all models is a pre-processed chat that concatenates the utterances of the help-seeker and removes non-Hebrew characters and URLs. (See \autoref{subsec:trunc} for a comparison with the case of also including the utterances of the counselor). For each bag of words based model (W2V, TF-IDF, HPL), the embeddings were scaled to [0,1] range and fed into a logistic regression model.

  \paragraph{Fine-tuned AlephBERT (FT-BERT)}  
We used  the publicly available model of AlephBERT  for text classification and fine tuned it on the labels in the Sahar dataset. 


\paragraph{SI-BERT}
 The SI-BERT PLM described in~\autoref{subsec:si-bert}.
 
 \paragraph{Expert-based SI Lexicon (SI-Lexicon)} 
 The expert-based SI lexicon that is described in~\autoref{ssec:lex}.
 
\paragraph{W2V embeddings + Logistic Regression  (W2V-LR)} Word To Vector (W2V)  is an algorithm which uses a neural network to map each word to a vector with a fixed size  \cite{mikolov2013efficient}. We trained a W2V model  on our corpus (embedding dimension=300, as used in the original paper \cite{mikolov2013efficient}).  The model was used to generate an   embedding of all words in the session.  
\paragraph{TF-IDF  + Logistic Regression  (TF-IDF-LR)}  
Term Frequency–Inverse Document Frequency (TF-IDF) is a term weighting scheme commonly used to represent textual documents as vectors \cite{sammut2010tf}.
Each session is vectorized in this manner.


    \paragraph{Hebrew Psychological Lexicon + Logistic Regression (HPL-LR)}
Each session is mapped to a vector of length 276 (number of lexicons in HPL). The $i$th element in the vector, is the number of occurrences of phrases in lexicon $i$ in a given session.

 We focus on two metrics to evaluate a model's performance: (a) ROC-AUC is the most commonly used metric in suicide detection tasks \cite{bernert2020artificial}. Its main advantage is   that it doesn't depend on class distribution in the dataset.
 \kg{add also about threshold insensitivity?}
(b) The F2 metric computes a weighted harmonic average between the precision and recall scores.
It assigns more than twice the weight (compared to the standard F1 metric) to the recall score which is sensitive to false-negative classifications. False negatives are critical in the SI detection task since missing people at risk has life-threatening consequences.


\section{Results}
In this section we report and discuss results from four perspectives: (i) Entire sessions results, (ii) Early detection, (iii) Results on different demographic groups, and (iv) Using the turn-taking structure vs. focusing on the utterances of the help-seeker.  
All results are reported using 5-fold cross validation. 
We keep the label imbalance unchanged in order to increase the potential use of the models for real time SI detection, discussing the False Positive-False Negative trade-off. \ab{probelm}
 The input to all models used in ~\autoref{subsec:gen_res} - ~\autoref{subsec:dem_res} consists of the concatenated utterances of the help-seeker in each session, while in \autoref{subsec:trunc} we also provide results for a turn-taking scenario.

\subsection{Entire Sessions Results}
\label{subsec:gen_res}

\begin{table*}
\centering
\small
  \begin{tabular}{rlllll}\toprule
    \textit{Model} & \textit{ROC-AUC[\%]} & \textit{F1[\%]} & \textit{F2[\%]}  & \textit{Precision[\%]} & \textit{Recall[\%]}\\ \midrule
    W2V-LR & $86(0.25)$ & $42(1.35)$ &$33(1.50)$&$75(1.86)$&$29(1.50)$\\
    TF-IDF-LR & $82(0.44)$ & $43(0.30)$ &$34(0.45)$ & $75(1.50)$&$30(0.50)$\\
    HPL-LR & $78(0.51)$ & $28(0.94)$ &$21(0.87)$&$66(2.53)$&$18(0.81)$\\
 
    SI-Lexicon  & $82(0.67)$ & $51(0.58)$ &$42(0.59)$&$\mathbf{78(1.54)}$&$38(0.60)$\\
    FT-BERT & $84(0.47)$ & $47(1.35)$ & $39(1.63)$&$70(1.61)$&$42(1.74)$\\
    SI-BERT & $87(0.37)$ & $55(1.21)$ &$49(1.54)$&$71(1.52)$&$45(1.69)$ \\
    Ensemble  & $\mathbf{91(0.45)^*}$ & $\mathbf{61(0.89)^*}$ &$\mathbf{55(1.32)^*}$  &$76(2.21)$&$\mathbf{51(1.59)^*}$
 \\ \bottomrule
  \end{tabular}
    \caption{SI classification results listing average performance with standard error in parenthesis. Bold highlights highest value, $^*$ marks a model is significantly better than the rest with $p<0.05$ under Wilcoxon signed rank test.}

    \label{tab:FT_Scores}
\end{table*}

\autoref{tab:FT_Scores} compares the different models using 
several metrics, including 
ROC-AUC, and F2, which  are commonly used  for settings that suffer from high class imbalance~\cite{forman2010apples}.

 As shown in the table, the Ensemble model significantly outperforms the other models in terms of F1, F2 and ROC-AUC metrics.  SI-BERT  outperforms FT-BERT, and the other baselines. The SI-Lexicon achieves the highest precision, slightly better that the Ensemble,  and does very well compared to the non-PLM approaches in the F1 and F2 metrics. However, it achieves only modest recall.  This reflects the  fact that the lexicon model was manually crafted by mental clinicians and tailored to detect explicit use of suicide ideation.  However, there still exist SI positive sessions that cannot be captured by a static list of phrases.

\begin{figure}[h]
    \centering
     \includegraphics[width=5cm]{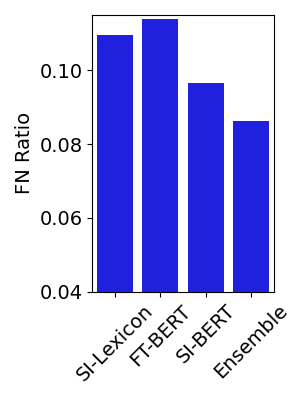}%
      \captionsetup{width=.48\textwidth}
     \caption{ False-negative  ratio (out of test-set size) for top-performing SI detection approaches.}
    \label{fig:conf_mat}
\end{figure}

The Ensemble model achieves a significant improvement in recall compared to the models it is comprised of, with only a slight decrease in precision performance. We wish to stress that this trade-off is of major significance in this specific domain. To further illustrate this point, \autoref{fig:conf_mat} presents false-negative ratios (out of test-set size) for the top performing models.
 As shown in the figure, the   Ensemble model achieves 
the lowest false-negative ($8.6\%$) 
given that the positive  SI samples account for $17\%$ of the test data.
Most importantly, the Ensemble model reduces the  false negatives ratio from $10.96\%$  to $8.60\%$  ($11\%$ decrease).  In the field, such an improvement provides a meaningful contribution to suicide prevention, especially in early stages of the session, as we show in the following subsection.

   
    
\subsection{Early Detection}
\label{subsec:early_detection}
 
\begin{figure}[h]
    \centering
     \includegraphics[width=7.7cm]{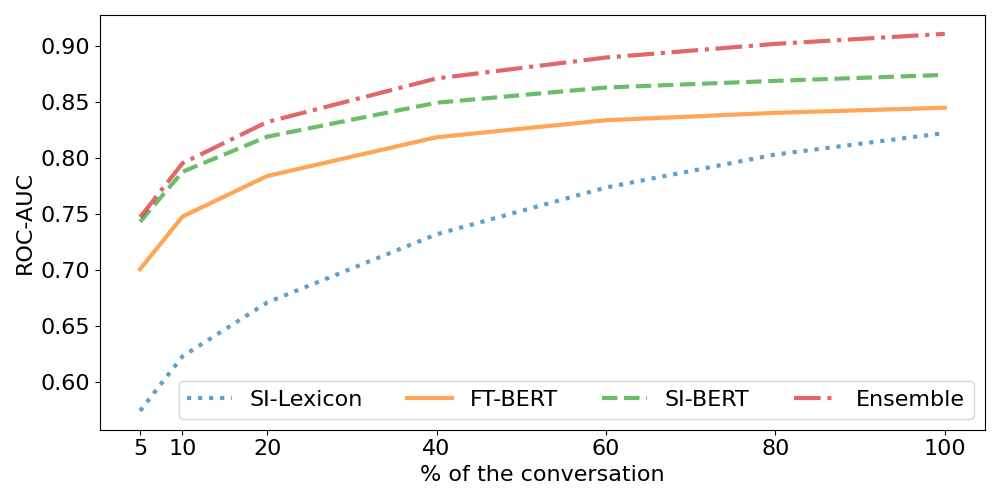}
      \captionsetup{width=.48\textwidth}
     \caption{Classification results for early detection of top-performing SI detection approaches.}
    \label{fig: temporal}
\end{figure}

Detecting at-risk individuals as early as possible during the session contributes to suicide prevention and  reduces the load on the volunteers.  To this end, \autoref{fig: temporal} shows the ROC-AUC performance of  the top-performing SI-detection approaches     when analyzing the first $\{5,10,20,40,60,80,100\}$ percent of the session (using 5-fold cross validation). As expected, all of the approaches improve as they process more information, with the Ensemble model constantly outperforming all of the other approaches.
 
 Two key findings that stand out from \autoref{fig: temporal} are: (a) There is a consistent  gap in performance between SI-BERT and FT-BERT, especially at early stages of the conversation.  (b) SI-Lexicon performs  poorly at an early stage of the conversation. We hypothesize that help seekers tend to be implicit, before allowing  themselves to express their suicidal tendency explicitly\footnote{Furthermore, the explicit expressions could be a response to the counselor sensing implicit cues before directing the conversation a certain way.}. Specifically, we  found that while 73\% of SI positive sessions contained an explicit SI phrase from the lexicon, only 38\% of the SI positive sessions contained an explicit SI phrase in the early 20\% of the session. This strengthens our conclusion that the lexicon model is insufficient for early detection of suicide risk.
 


 
\begin{table*}[t]
\small
\begin{center}{
   \begin{tabular}{ccccccccc}\toprule
  \small
    \textit{Age/Gender} & \textit{Samples} & \textit{$+$ Label}&  \textit{ROC-AUC} & \textit{F2} &\textit{Tokens}&\textit{Types}&\textit{OOV Tokens} &\textit{OOV Types}\\ \midrule
    10-17 & $4,179 (23\%) $&$15\%$ & $90$& $52$ &1.1M&54K&$6\%$&$6\%$ \\
    18-30 & $9,066  (52\%)$& $19\%$  & $90$&  $55$ &2.8M&141K&$8\%$ &$7\%$\\
    31-64 & $4,164 (24\%)$&$17\%$   & $91$&  $56$ &1.2M&107K&$9\%$&$6\%$ \\
    65+ & $145 (<1\%)$  &$12\%$& $93$&  $58$ &29K&8K&$12\%$&$4\%$ \\
    \hline \\
    Female & $12,074$ &$18$& $90$ & $55$&$3.6M$&$120K$ &$8\%$&$8\%$\\
    Male & $5,343$ &$17$& $91$ & $55$ &$1.6M$&$83K$ &$8\%$&$7\%$\\
 \\ \bottomrule
  \end{tabular}
    \caption{Ensemble model performance evaluation for subgroups of different age and gender.}
    \label{tab:dem_subgroups}
 }\end{center}
\end{table*}

\subsection{Demographic Analysis}
\label{subsec:dem_res}
Suicide risk, technological proficiency and linguistic norms vary across demographics.  Therefore, we evaluate the performance of our model over different demographics.

\paragraph{Age} The Ensemble model consistently outperforms all other models across all age groups.
Results of the Ensemble for different age  groups are presented in \autoref{tab:dem_subgroups} (Top), along with descriptive statistics, highlighting the differences between the sub-corpora in terms of size, types, tokens and OOV types and tokens. 

The different linguistic norms each age group exhibits is captured in \autoref{tab:unique_score} through the relative size of each group's unique vocabulary. For example, while the sub-corpora 10-17 and 31-64 are a similar share of the data (23\% \& 24\%, see \autoref{tab:dem_subgroups}) and a similar label break down (15\% \& 17\%), 39\% of the tokens\ab{types?} used by the  the 10-17 help-seekers are not used by help-seekers 31-64 years of age. Similarly, 54\% of the tokens\ab{types?} used by the 31-64 group are not used by the individuals on the 10-17 group. These trends are even more pronounced if one considers only OOV types. 

Given the large variance in vocabularies between groups, the consistency of our results further demonstrates the robustness of our model. 

\begin{table}
\centering
\small
  \begin{tabular}{ccccc}\toprule
     & \textit{10-17} & \textit{18-30} & \textit{31-64} & \textit{65+} \\ \midrule
    10-17 & $-$ & $0.28$ &$0.39$&$0.86$\\
    18-30 & $0.60$ & $-$ &$0.55$&$0.91$\\
    31-64 & $0.54$ & $0.37$ &$-$&$0.89$\\
    65+ & $0.21$ & $0.14$ &$0.16$&$-$\\
 \\ \bottomrule
  \end{tabular}
    \caption{Percentage of the number of unique \ab{changed from tokens to tokens} tokens of each age group with respect to other age groups.}
    \label{tab:unique_score}
\end{table}

\paragraph{Gender} Examining the two main gender categories\footnote{The gender the help-seeker identifies with is implicitly self disclosed since Hebrew is a gendered language and first-person verbs often takes different suffix according to the speaker gender (see examples in \autoref{footnote:gender_cases}).}
we verify that the model achieves similar performance for both genders (see \autoref{tab:dem_subgroups}, Bottom). This result is far from trivial for two reasons: (a) Hebrew is a heavily gendered language (e.g., \cite{vainapel2015dark}). This means
most (non-past tense) verbs and adjectives have different morphological inflection depending on the speaker's gender\footnote{For example, consider the Hebrew inflections of the adjective `lonely': `boded' (M) vs. `bodeda' (F), or the verb `going (to)': `holex' (M) vs. `holexet' (F).\label{footnote:gender_cases}}, and (b) The number of samples for each gender varies greatly, with female individuals making a vast majority of help seekers.

\subsection{Speaker and Text Truncation} \label{subsec:trunc}
One major limitation of the BERT architecture is the constraint it enforces on length of the input. Consequently, most of the sessions cannot be processed fully. A straight forward way to tackle this constraint is to feed the model only part of the session -- exhausting the 512-tokens buffer size. We considered three alternative protocols: (i) using utterances of both help-seekers and the counselor, (ii) using only utterances made by the help-seeker, and (iii) using only utterances made by the counselor. The latter protocol is used for comparative reason (also assuming that the responses of the counselor bear relevant signal). 
In each of these three settings, we considered two options (a) using the first 512 tokens (``keep head''), and (b) using the 512 trailing tokens (``keep tail'').

\begin{figure}
    \centering
     \includegraphics[width=0.48\textwidth]{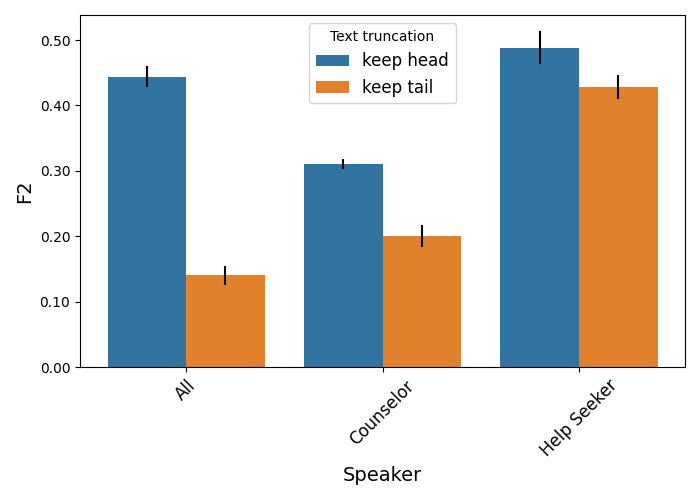}
      \captionsetup{width=.48\textwidth}
     \caption{SI-BERT F2 performance for different text truncation methods and speaker text. Error bars mark standard error.}
    \label{fig: head_or_tail}
\end{figure}

Results of the Ensemble for each of the six settings are presented in \autoref{fig: head_or_tail}. The best performance is obtained using the head of the the utterances made by the help-seeker. This result, together with the ability to perform relatively well in  early stages (\autoref{subsec:early_detection}) are encouraging, given the high stakes of the task and the limited resources (personnel) available to the emergency services. 

\section{Discussion}

The results in the previous section demonstrate the need for adequate adaptation of PLMs into a specific domain with  obvious challenges in processing low-resource languages.
Domain-tailored lexicons serve as strong baselines, with high precision and competitive F1 and F2 scores but they fall short in terms of recall and AUC. 
As PLMs tend to capture more nuanced expressions of SI --  combining both approaches and careful fine-tuning improved the ability to detect SI early on in the chat. 

We conclude this work discussing three illustrative examples (see \autoref{tab:examples}) and reflecting on a few limitations 
of our approach.  

\begin{table}[h]
\centering
\begin{tabular}{ m{.2cm}m{6.7cm} } 
 \toprule
    I & \emph{I don't want to die}\\
    \hline
    II & \emph{I feel like life is too much for me} \\
    \hline
    III & \emph{I had a spare time and bad thoughts, I decided to take a couple of sleeping pills and take a nap. I slept all day and now I'm dizzy.}\\
 \bottomrule
    \end{tabular}
\caption{Three illustrative utterances (translated from Hebrew). Note that these utterance are presented without a conversational context.}
\label{tab:examples}
\end{table}
 
In utterance I (\autoref{tab:examples}), the speaker explicitly rejects a suicidal intent, however, both the lexicon and SI-BERT (and the Ensemble, of course) classify this utterance as positive ideation. While the lexical approach matches the token \emph{die} and wrongfully ignores the negation, our collaborators -- clinical psychologists with suicide detection as their research focus -- approve the classification, citing  psychological studies on the distinction between suicide ideation and intent \cite{bagley1975suicidal,beck1979assessment,mcauliffe2002suicidal}.

While it does not match any of the lexical items in the predefined lexicons, the second utterance (\autoref{tab:examples}, II) is a classic example of SI. SI-BERT and the Ensemble (as well as the online counselor) correctly label it a positive SI, demonstrating the benefits of the domain-specific contextual model.

The third utterance is not considered by experts to exhibit SI. Indeed, SI-BERT (and the Ensemble) correctly classifies it as a negative example.  On the other hand,  FT-BERT assigns it a positive label. We hypothesize that the combination of ``bad thoughts'' and ``(a couple of) sleeping pills'' triggered the naive FT-BERT, while SI-BERT better captures the nuanced context. This example further demonstrates the benefits of fine-tuning the vanilla AlephBERT not only for the classification task but also fine-tuning the language model on the domain-specific data through a masked LM task.

We end this section briefly mentioning some limitations of the approach. First, our approach does not explicitly account for the discourse structure of the sessions. It may be that encoding the full conversational context  may improve performance. Second, the Ensemble approach relies on a hand-crafted lexicon requiring considerable human effort. 
Many psychological lexicons already exist in other languages and have played a considerable role in prior SI research, see  \cite{lee2020cross}. Investing further effort in lexicon creation may have further reduced the false negative rate.  
Third, the lack of a benchmark dataset for the SI detection task makes it difficult to compare with  prior work and approaches.

\section{Conclusion and Future Work}

Accurate and early detection of users' suicide risk in text-based counseling services is essential to ensure that at-risk individuals are given timely and proper treatment.  This paper provides an automatic approach to risk detection from chats in Hebrew, a low-resource and morpholigically rich language. 
  Our approach adapted a generic Hebrew language model by (i) adding out-of-vocabulary tokens, and (ii) performing additional pre-training of the LM on a masked language modeling task over the specific domain. Finally, we fine-tuned the model for the suicide risk detection task. We combined this model with a lexicon of hand crafted suicide ideation phrases that were vetted by experts.  Our Ensemble model outperformed several competitive approaches, including  a generic language model and the stand-alone lexicon. Our model performed consistently well for different demographics (age, gender). 
  These encouraging results suggest the model can be deployed successfully, providing the much needed support to volunteers and health-care professionals in their mission of reducing suicide rates.
  
In future work, we wish to integrate the discursive structure into the model, and include latent information about the cognitive state of the help-seeker. 


\bibliography{anthology,custom}
\bibliographystyle{acl_natbib}



\end{document}